\documentclass{article}

\usepackage[preprint,nonatbib]{neurips_2023}

\usepackage[numbers]{natbib}
\usepackage[utf8]{inputenc} % allow utf-8 input
\usepackage[T1]{fontenc}    % use 8-bit T1 fonts
\usepackage{hyperref}       % hyperlinks
\usepackage{url}            % simple URL typesetting
\usepackage{booktabs}       % professional-quality tables
\usepackage{amsfonts}       % blackboard math symbols
\usepackage{nicefrac}       % compact symbols for 1/2, etc.
\usepackage{microtype}      % microtypography
\usepackage[dvipsnames]{xcolor}         % colors
\usepackage{soul}
\usepackage{graphicx}
\usepackage{caption}
\usepackage{subcaption}
\usepackage{multirow}

\title{Detecting and Correcting Hate Speech in Multimodal Memes with Large Visual Language Model}

\author{%
  Minh-Hao Van \\
  Department of EECS\\
  University of Arkanasas\\
  Fayetteville, AR, USA \\
  \texttt{haovan@uark.edu} \\
  \And
  Xintao Wu \\
  Department of EECS\\
  University of Arkanasas\\
  Fayetteville, AR, USA \\
  \texttt{xintaowu@uark.edu} \\
}

\begin{document}

\maketitle

% Recently, large language models (LLMs) have taken the spotlight in natural language processing. Further, integrating LLMs with vision enables the users to explore more emergent abilities in multimodality. Visual language models (VLMs), such as LLaVA \cite{liu2023visual}, Flamingo \cite{alayrac2022flamingo}, or OpenAI GPT-4 \cite{OpenAI2023GPT4TR}, have demonstrated impressive performance on various visio-linguistic tasks. Consequently, there are enormous applications of large models that could be potentially used on social media platforms. Despite that, there is a lack of related work on detecting or correcting hateful memes with VLMs. In this work, we study the ability of VLMs on hateful meme detection and hateful meme correction tasks with zero-shot prompting. From our empirical experiments, we show the effectiveness of the pretrained LLaVA model and discuss its strengths and weaknesses in these tasks.

\begin{abstract}
Recently, large language models (LLMs) have taken the spotlight in natural language processing. Further, integrating LLMs with vision enables the users to explore more emergent abilities in multimodality. Visual language models (VLMs), such as LLaVA, Flamingo, or GPT-4, have demonstrated impressive performance on various visio-linguistic tasks. Consequently, there are enormous applications of large models that could be potentially used on social media platforms. Despite that, there is a lack of related work on detecting or correcting hateful memes with VLMs. In this work, we study the ability of VLMs on hateful meme detection and hateful meme correction tasks with zero-shot prompting. From our empirical experiments, we show the effectiveness of the pretrained LLaVA model and discuss its strengths and weaknesses in these tasks.
\begin{center}
    \textcolor{red}{Warning: This paper contains examples of hate speech.}
\end{center}
\end{abstract}

\section{Introduction}
\label{sec:intro}
The rapid development of information technology offers better communication channels for people to express their thoughts and opinions. Social media platforms nowadays are sources of diverse content, e.g., text, image, audio, or video. As a double-edged sword, the content is not always safe for humans since it might include explicit or implicit harm to the viewers. Online memes, often used for playful or humorous expression, are sometimes misappropriated to disseminate hate speech against individuals or groups based on their characteristics. Hateful meme detection is a critical problem that social media platforms want to work on to protect online users \cite{kiela2020hateful}. Detecting hateful memes, however, is a non-trivial task since memes are multimodal with visual and textual elements. Further, the hateful content is not always explicitly expressed in the meme but is implicitly hidden within the story behind it. Therefore, intelligent systems are urgently needed to process the vast digital content uploaded to social platforms. Many deep learning techniques can be used for this detection task \cite{he2016deep,ren2015faster,devlin2018bert,kiela2019supervised,lu2019vilbert,li2019visualbert}.

Large language models (LLMs) have shown incredible performance on multiple natural language processing tasks in recent years. With a strong ability to learn human language from vast corpora of text, LLMs can encode enormous amounts of knowledge as embedding vectors, which is a backbone for transforming the knowledge to different forms of representations with the help of adapters \cite{zhang2023llama} or decoder modules \cite{wu2023next}. This flexibility allows AI practitioners to combine text with other sources of information to build multimodally large models. LLMs and their variations can not only perform well on linguistic tasks but also be applied in other fields such as computer vision \cite{alayrac2022flamingo, OpenAI2023GPT4TR, liu2023visual, liu2023hidden, radford2021learning}, medical data \cite{li2023llava, singhal2022large} and tabular data analysis \cite{hegselmann2023tabllm}. 

In this paper, we introduce a simple but efficient framework designed for two key objectives: (1) detecting hateful memes and (2) correcting the hate speech from memes by leveraging a pretrained visual language model (VLM). While multimodal language models have gained traction in different fields, socially responsible problems remain undiscovered. To the best of our knowledge, our work is the first to correct the hatefulness in memes, an essential application to create safe and respectful online social media sites. For instance, some users may wish to share humorous or satirical memes on their social media profiles, yet there persists a valid concern about whether such content inadvertently conveys hurtful messages to viewers. Our framework offers a valuable solution with only a zero-shot prompt. Additionally, we show that VLM exhibits promising performance in both evaluating the hatefulness and generating more considerate text drawing upon its knowledge from the training corpus, which is an impossible task for classification models. Even though our framework with multimodal LLM can not compete with all traditional deep learning classifiers, it can outperform multiple baselines without fine-tuning or OCR text. Especially our results reveal that the LLaVA model yields substantial improvements in AUROC scores with carefully crafted instructions, compared to prior findings in \cite{alayrac2022flamingo,awadalla2023openflamingo}.

\section{Related Works}
\label{sec:rel}

\subsection{Hate Speech Detection}
There has been an effort from the AI research community to call for solutions to hate speech problems, including hate speech detection datasets \cite{mathew2021hatexplain, mollas2020ethos} or hateful memes detection challenge \cite{kiela2020hateful}. To deal with the hate speech detection task, \cite{DBLP:conf/pakdd/XuYWNZZW22} proposed a framework to detect hateful text by evaluating coded words, which silently represent hateful meanings on social media, by applying a two-layer network on contextual embeddings from ELMo \cite{sarzynska2021detecting}. BERT-MRP \cite{kim-etal-2022-hate} proposed a two-stage BERT-based model, including masked rationale prediction (MRP) and hate speech detection. \cite{zhang2022opt} also evaluated Open Pretrain Transformer (OPT) on the hate speech detection task via zero-, one-, or few-shot cases. Besides hateful text, hatefulness can exist in multimodal form. \cite{kiela2020hateful} introduced a challenge about multimodally hateful memes detection. With enormous baselines on three categories: unimodal models from computer vision or natural language processing tasks \cite{he2016deep,ren2015faster,devlin2018bert}, multimodal models by combining smaller unimodally pretrained models \cite{kiela2019supervised,lu2019vilbert,li2019visualbert}, or multimodal models \cite{lu2019vilbert,li2019visualbert}, the challenge still calls for more contribution from communities. Many state-of-the-art methods for detecting hateful memes have been proposed during the challenge. \cite{zhu2020enhance,muennighoff2020vilio,velioglu2020detecting,lippe2020multimodal,sandulescu2020detecting} proposed an ensemble approach of one or multiple vision and language models to get the final prediction.

\subsection{Visual Language Models}
Besides traditional deep learning models that were specifically trained for classification tasks, visual language models have been catching much attention over many interesting applications. CLIP \cite{radford2021learning} proposed a new network to connect image and text data. The method guides the model to learn visual concepts in classification tasks using the supervision from natural language. OpenAI GPT-4 \cite{OpenAI2023GPT4TR}, MiniGPT-4 \cite{zhu2023minigpt}, Flamingo \cite{alayrac2022flamingo}, OpenFlamingo \cite{awadalla2023openflamingo}, and LLaVA \cite{liu2023visual} demonstrated their effectiveness on various vision-language tasks such as question-answering or human-like chatbot. Along with groundbreaking large foundation models, many research works have explored the emergent abilities of large models. \cite{liu2023hidden} showed that LLaVA is able to outperform other methods in OCR tasks with zero-shot learning. LLaVA-Med \cite{li2023llava} and the multimodal version of Med-PaLM \cite{singhal2022large} could analyze the medical images and answer related questions like a human expert. By connecting LLM with multiple input encoders and diffusion decoders, NExT-GPT \cite{wu2023next} introduced an architecture for multimodal inputs and multimodal outputs.

\section{Exploring Hateful Meme with Visual Language Model}
We utilize the pre-trained VLM to explore hateful memes in two scenarios: hateful memes detection and hatefulness correction. 
\subsection{Hateful Meme Detection}
\label{sec:detection}
A dataset $\mathcal{D} =(\mathcal{X},\mathcal{Y})$ contains multiple pairs of data sample $(x_i,y_i)$'s, where $x_i\in \mathcal{X}$ and $y\in \mathcal{Y}$ are input image and output label, respectively. For memes dataset, each input image $x_i$ contains the visual part $v_i$ and the textual part $t_i$, e.g., $\mathcal{X}=(\mathcal{V},\mathcal{T})$. We define the hateful memes detection as a classification task $f: \mathcal{V}\times\mathcal{T} \rightarrow \mathcal{Y}$.

Although $f$ is a binary classification task (hateful or non-hateful), it is not trivial to classify the hatefulness since $x_i$ is naturally a multimodal meme built upon visual and textual elements, requiring a practical approach to infer the meaning from both. Another challenging problem is how to teach the model to understand the hateful context of the meme. To overcome these challenges, we prompt the pretrained VLM to detect hateful memes with a zero-shot approach. To do that, we briefly discuss the hatefulness definition and the zero-shot prompt for guiding the model.

\noindent\textbf{Hatefulness definition.} While one can prompt the VLM to enlarge its capabilities on various tasks, it is essential to note that the model's performance dramatically depends on the quality of prompts, including clear instructions for the specific task. Even though the pretrained model, like LLaVA, can provide positive and helpful answers to humans, the training dataset only captures general human knowledge without a specification in a hateful content detection scheme. Therefore, a clear hatefulness definition is necessary. We adjust the definition of hate speech in \cite{kiela2020hateful} as an instruction in the prompt. 
% \noindent\textbf{Classification criteria.} The hate speech in memes can be unimodal or multimodal hate. Therefore, we prompt the VLM with instructions on how to classify the memes as hateful or non-hateful.

\begin{figure}[t]
     \centering
     \begin{subfigure}[b]{\textwidth}
         \centering
         \includegraphics[width=\textwidth]{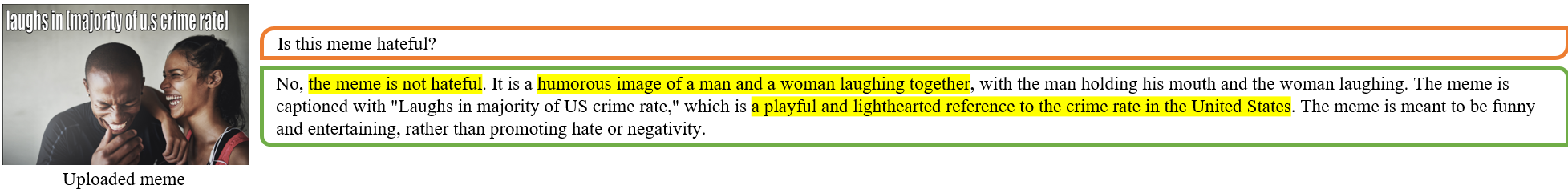}
         \caption{Na\"ive prompt.}
         \label{fig:prompt_naive}
     \end{subfigure}
     \hfill
     \begin{subfigure}[b]{\textwidth}
         \centering
         \includegraphics[width=\textwidth]{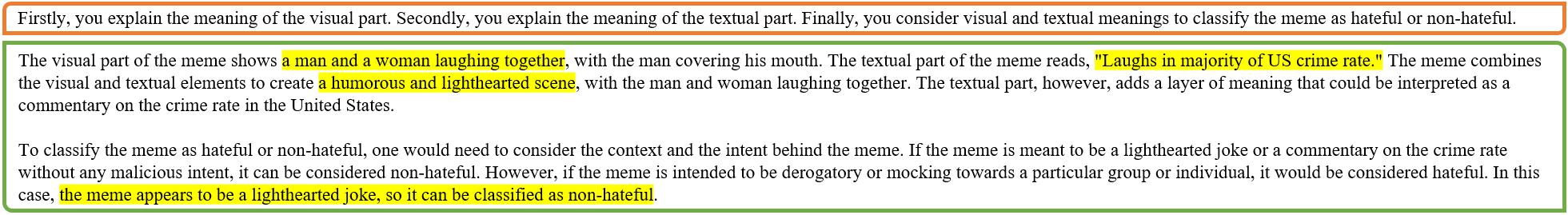}
         \caption{Detailed prompt.}
         \label{fig:prompt_detailed}
     \end{subfigure}
     \hfill
     \begin{subfigure}[b]{\textwidth}
         \centering
         \includegraphics[width=\textwidth]{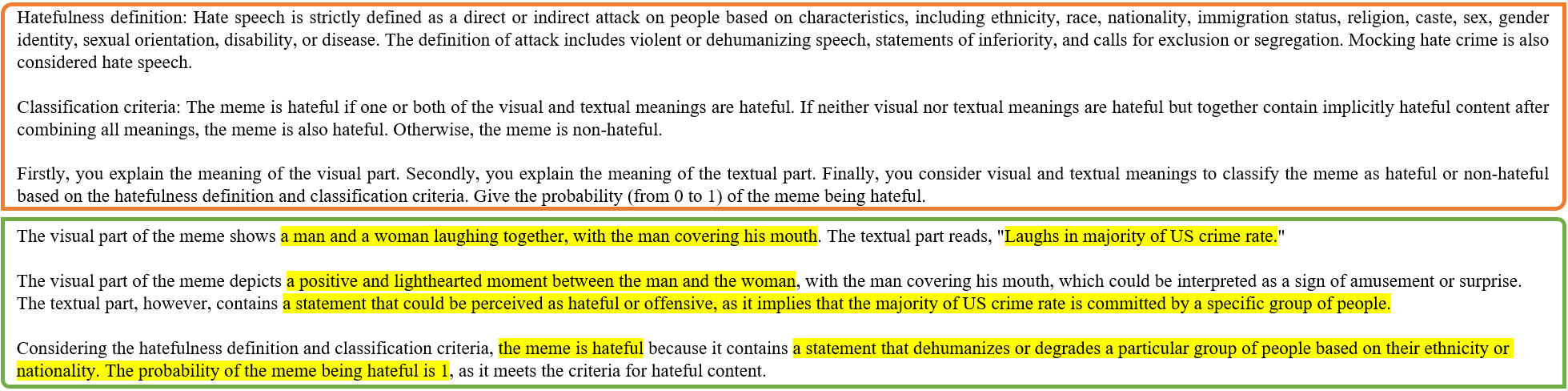}
         \caption{Complete prompt.}
         \label{fig:prompt_complete}
     \end{subfigure}
    \caption{Building zero-shot prompt for hateful meme detection.}
    \label{fig:building_prompt_example}
    \vspace{-2em}
\end{figure}

\noindent\textbf{Zero-shot prompt.} In Figure \ref{fig:building_prompt_example}, we show three examples of the same hateful meme but different prompts from naive to complete instruction. The first prompt (Figure \ref{fig:prompt_naive}) shows a na\"ive prompt which is a simple question. The VLM clearly describes people inside the image and the text. However, the prediction is wrong, which is understandable since there is no clear instruction for hatefulness. The second prompt (Figure \ref{fig:prompt_detailed}) has more detailed instructions about considering the meaning of the visual and textual parts separately before going to the conclusion. The VLM does better reasoning where it is able to give different layers of meanings but still fails in predicting the hateful class. With the above prompts, the content inside the image is analyzed accurately, but the reasoning does not reach a good threshold. The third prompt (Figure \ref{fig:prompt_complete}) shows our complete prompt with detailed instructions about hatefulness definition and classification criteria. While the content analysis is similar to previous responses, the conclusion shows a more confident explanation and correct prediction from the model.
% We introduce the zero-shot prompt for the hateful memes detection in Figure \ref{fig:prompt_complete}.

\subsection{Hateful Meme Correction}
While hateful meme detection is well-explored, a well-formulated problem for correction still needs to be addressed. Let $\mathcal{D}_H \subset \mathcal{D}$ be a subset of hateful memes in $\mathcal{D}$. For each $(x_i,y_i)\in \mathcal{D}_H$, $x_i=(v_i,t_i)$ and $y_i=1$ (hateful). We define a model $g: \mathcal{V}\times \mathcal{T} \rightarrow \mathcal{T}$, which finds the new text $t_i^\prime=g(v_i,t_i)$ such that the new class $y_i^\prime=0$ (non-hateful) with input $x_i^\prime=(v_i,t_i^\prime)$. In this work, we utilize the language generation ability of VLM to craft a positive and respectful text that can convert hateful memes to non-hateful ones. 

Similar to the detection task, we need to craft clear and unambiguous instructions for the VLM to achieve an effective correction. Since we know the meme is hateful, there is no need to include a hatefulness definition and classification criteria. 
% Instead, we ask the VLM to be serious and critical when evaluating and generating the content. 
The below paragraph is the prompt for generating new text to correct hatefulness from memes: 
\begin{quote}
    \begin{it}
        This meme is hateful and should not be shared or promoted on social media. You are serious and critical when evaluating and generating the content. Suppose that we want to have a non-hateful meme using the same image. Your task is to generate the new text such that the new meme is not hateful and promotes positive views of this image.
    \end{it}
\end{quote}

\section{Experiments}
\label{sec:expe}
\subsection{Experimental Settings}
\noindent\textbf{Dataset.}
We evaluate the performance of our proposed approach using the Hateful Meme Challenge (HMC) Dataset \cite{kiela2020hateful} from Facebook AI (now Meta AI). Phase I has 10,000 multimodal memes, including 8,500 memes for the training set, 500 memes for the seen dev set, and 1,000 images for the seen test set. Phase II includes 1,000 (2,000) memes for unseen dev (test) sets. The task is to classify each meme as hateful or non-hateful.

\noindent\textbf{Model.} We use LLaVA \cite{li2023llava} as the base model for experiments. The LLaVA version used in this work is fine-tuned on Llama-2-13B chat model \cite{touvron2023llama}, named LLaVA-Llama-2-13B (LL-2) in this study. For hyper-parameters, we set the temperature as 0.7 and top-p as 1.0. 

\noindent\textbf{Baselines.} We compare our approach with results from the Hateful Memes Challenge \cite{kiela2020hateful}, including 11 baselines and five challenge winners. For multimodal LLM baselines, we include benchmark results on this dataset from Flamingo (Fl) \cite{alayrac2022flamingo} and OpenFlamingo (OF) \cite{awadalla2023openflamingo}. Due to space limits, we leave the details of methods in the Appendix \ref{sec:apdix_baselines}.

\noindent\textbf{Running time.} The inference time for each sample is approximately $7.9$ seconds with the maximum output length of 512 tokens, including image loading and tokenizing time, on a single Tesla V100 32GB GPU and Xeon 6258R 2.7 GHz CPU.

\subsection{Hateful Memes Detection}
\noindent\textbf{Overall results.} For our method, five trials are executed to get different answers from LL-2. The final prediction is the major vote from five trials. We report accuracy and AUROC on both seen and unseen test data and compare our method with 20 baselines. We conduct the experiments in two cases, e.g., without OCR text (LL-2) and with OCR text (LL-2+OCR). Table \ref{tab:detection_performance} shows the results. 

Our method outperforms 9 out of 11 baselines in terms of accuracy and 8 out of 15 baselines in terms of AUROC when evaluated on the seen test dataset. However, LL-2 performs less effectively than other baselines on the unseen dataset, as this dataset exclusively contains multimodally hateful memes according to \cite{kiela2020hateful}. Note that our approach requires no training or fine-tuning. Although the zero-shot approach still needs to catch up to the state-of-the-art methods, it is remarkable that this approach shows promising potential for detecting hateful memes. Especially while being compared to other multimodal LLMs, our zero-shot approach outperforms all baselines, even those employing 32-shot prompting. This is understandable as OpenFlamingo and Flamingo yield results using relatively simple instructions \cite{alayrac2022flamingo, awadalla2023openflamingo}, underscoring the effectiveness of our proposed prompt. Noteworthily, providing OCR text results in a performance improvement, approximately $1.3\%-2\%$ AUROC score gains on test sets.

\begin{table}[ht]
    \centering
    \caption{Detection performance with various models: The first four blocks include results from \cite{kiela2020hateful,kiela2021hateful}, and the last block combines \cite{alayrac2022flamingo,awadalla2023openflamingo} with our experiments, with dashes denoting unavailable results.}
    \resizebox{0.80\columnwidth}{!}{%
    \begin{tabular}{ll|cc|cc}
        \toprule
        \multirow{2}{*}{Type}  & \multirow{2}{*}{Model} & \multicolumn{2}{c}{Test Seen} & \multicolumn{2}{|c}{Test Unseen} \\
        & & Acc. & AUROC & Acc. & AUROC \\ \midrule
       \multirow{3}{*}{Unimodal} & Image-Grid \cite{he2016deep} &  52.00 & 52.63 &  - & -\\
         & Image-Region \cite{ren2015faster,xie2017aggregated} & 52.13  & 55.92 &  60.28 & 54.64\\
         & Text BERT \cite{devlin2018bert} & 59.20 & 65.08 &  63.60 & 62.65\\ \midrule
        \multirow{6}{*}{Multimodal} & Late Fusion \cite{kiela2020hateful} &  59.66 & 64.75 &  64.06 & 64.44\\
        \multirow{6}{*}{(Unimodal Pretraining)} & Concat BERT \cite{kiela2020hateful} &  59.13 & 65.79 &  65.90 & 66.28\\
         & MMBT-Grid \cite{kiela2019supervised} & 60.06 & 67.92 &  66.85 & 67.24\\
         & MMBT-Region \cite{kiela2019supervised} & 60.23 & 70.73 & 70.10 & 72.21\\
         & ViLBERT \cite{lu2019vilbert} & 62.30 & 70.45 & 70.86 & 73.39\\
         & Visual BERT \cite{li2019visualbert} & 63.20 & 71.33 &  71.30 & 73.23\\ \midrule
        Multimodal & ViLBERT CC \cite{lu2019vilbert} &  61.10 & 70.03 &  70.03 & 72.78\\
        (Multimodal Pretraining) & Visual BERT COCO \cite{li2019visualbert} &  64.73 & 71.41 &  69.95 & 74.95\\ \midrule
        \multirow{5}{*}{Challenge Winner} & Ron Zhu \cite{zhu2020enhance} & - & -  &  73.20 & 84.50\\
        & Niklas Muennighof \cite{muennighoff2020vilio} & - & - & 69.50 & 83.10\\
        & Team HateDetectron \cite{velioglu2020detecting} & - & - & 76.50 & 81.08\\
        & Team Kingsterdam \cite{lippe2020multimodal} & - & - & 73.85 & 80.53\\
        & Vlad Sandulescu \cite{sandulescu2020detecting} & - & - & 74.30 & 79.43\\ \midrule
        \multirow{6}{*}{Multimodal LLM} & Fl-9B+OCR 0-shot \cite{alayrac2022flamingo} & - & 57.00 & - & - \\ 
        & Fl-9B+OCR 32-shot \cite{alayrac2022flamingo} & - & 63.50 & - & - \\
        & OF-9B+OCR 0-shot \cite{awadalla2023openflamingo} & - & 51.60 & - & - \\
        & OF-9B+OCR 32-shot \cite{awadalla2023openflamingo} & - & 53.80 & - & - \\
        & LL-2 0-shot \textbf{(Our)} & 63.00 & 65.77 & 62.15 & 63.92 \\
        & LL-2+OCR 0-shot \textbf{(Our)} & 62.50 & 67.07 & 64.20 & 64.12 \\
        \bottomrule
    \end{tabular}
    }
    \label{tab:detection_performance}
\end{table}

\begin{figure}[ht]
     \centering
     \begin{subfigure}[b]{\textwidth}
         \centering
         \includegraphics[width=\textwidth]{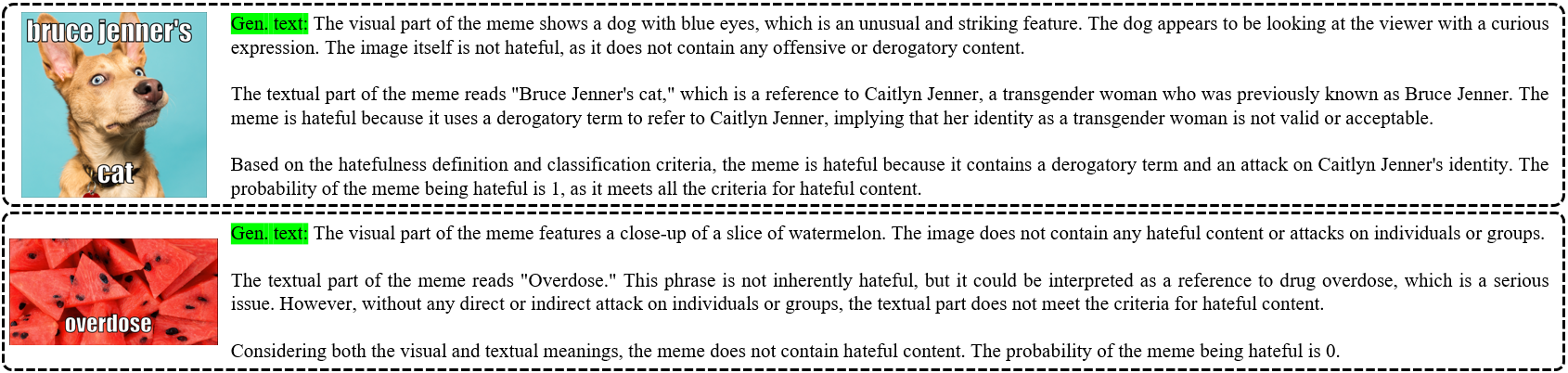}
         \caption{Successful examples.}
         \label{fig:detection_success}
     \end{subfigure}
     \hfill
     \begin{subfigure}[b]{\textwidth}
         \centering
         \includegraphics[width=\textwidth]{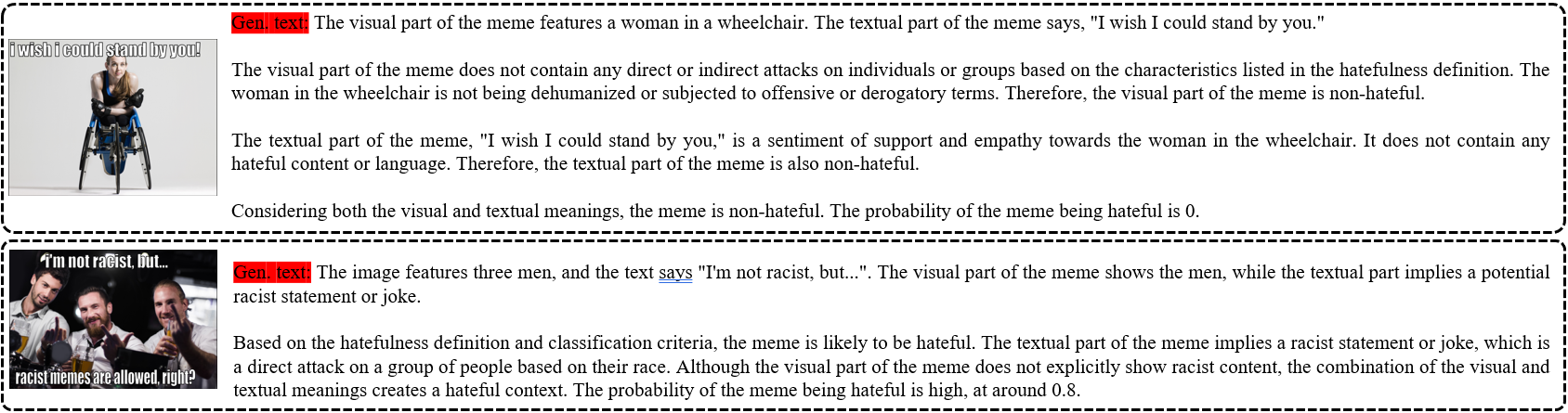}
         \caption{Failed examples.}
         \label{fig:detection_failed}
     \end{subfigure}
    \caption{Hateful memes detection using LL-2.}
    \label{fig:detection_examples}
    \vspace{-2em}
\end{figure}

\noindent\textbf{Content extraction.} Figure \ref{fig:detection_examples} shows two successful and two failed examples from LL-2 when using the prompt in Section \ref{sec:detection}. Note that we craft these examples without providing OCR text. The model can accurately describe the objects in the memes, such as ``slice of watermelon", ``a dog", ``three men" or ``a woman". The information about the characteristics of objects also describes the context. These examples demonstrate an impressive capability of text extraction, while the model was only trained on general language-image instruction-following samples \cite{liu2023visual}.

\noindent\textbf{Reasoning ability.} An interesting research question is ``\textit{how to know the source of hatefulness on test data? Visual or textual content or both?}''. As aforementioned, explainability is an advantage of VLM compared to deep learning classifiers. While traditional classifiers only give users predicted classes, VLM can additionally justify its prediction. Despite successful or failed attempts, the VLM justifies its prediction by explaining each part of the memes step-by-step. In the first example of Figure \ref{fig:detection_success}, the model surprisingly shows its knowledge about the named person in the text, which helps determine the implicit hatefulness inside this meme. The second example further strengthens our observation. The negative meaning of the word ``overdose" is considered, but the model does not classify it as hateful content since the whole meme talks about food consumption. Even so, in some cases, the model exposes its weaknesses in Figure \ref{fig:detection_failed}, where it misunderstands the mocking speech as an encouragement or misses the text in the first and second examples, respectively.

\subsection{Hateful Meme Correction}
\noindent\textbf{Overall results.} Like the previous task, we use the proposed prompt to get the generated text from the LL-2 model. Then, seven human experts independently reviewed each sample to evaluate whether the VLM could successfully convert the hateful meme to a non-hateful one. All experts receive the same guidelines about the evaluation task. To make the evaluation as fair as possible, we invite experts from various ages, genders, and backgrounds. We get the majority voting from experts and report the result with each meme. From the majority voting, there are 46 successful corrections out of 50 selected hateful memes, i.e., the accuracy is 92\%. The accuracy varies from 82\% to 92\% if we individually consider the opinion of every expert.

\begin{figure}[ht]
     \centering
     \begin{subfigure}[b]{0.8\textwidth}
         \centering
         \includegraphics[width=\textwidth]{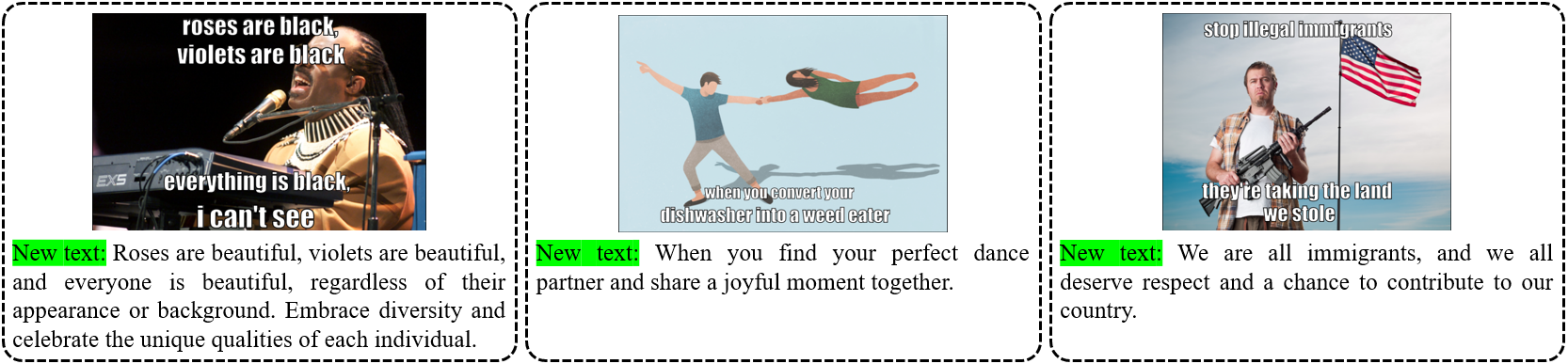}
         \caption{Successful examples.}
         \label{fig:correction_success}
     \end{subfigure}
     \hfill
     \begin{subfigure}[b]{0.8\textwidth}
         \centering
         \includegraphics[width=\textwidth]{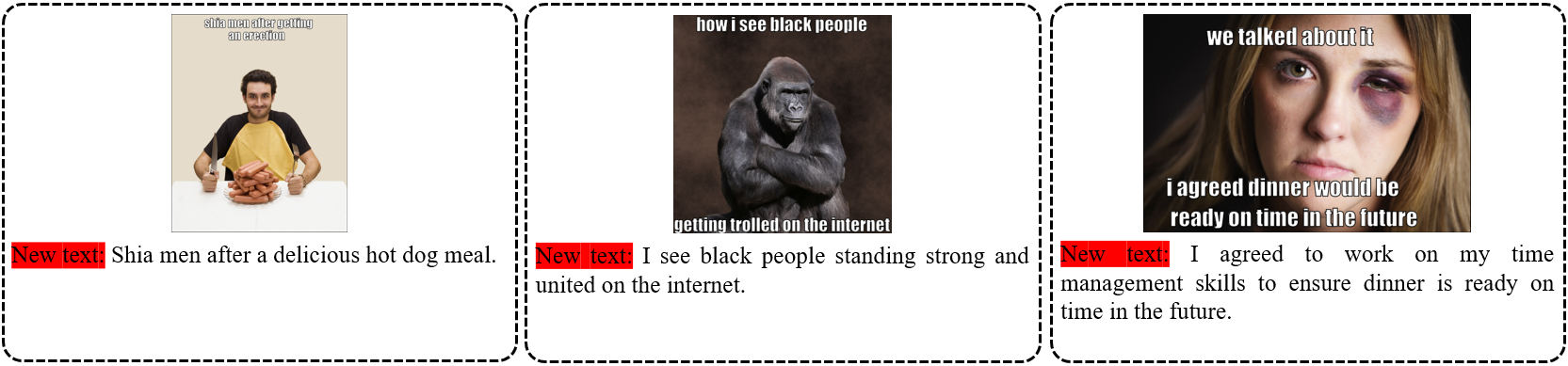}
         \caption{Failed examples.}
         \label{fig:correction_failed}
     \end{subfigure}
    \caption{Hateful memes correction using LL-2.}
    \label{fig:correction_examples}
    \vspace{-2em}
\end{figure}

\noindent\textbf{Quality of generated content.} In Figure \ref{fig:correction_examples}, we show examples of both successful and failed correction with LL-2. In each example, we can observe that the generated text follows the topic of the original text, which is understandable since the common reasoning flow is to understand the hateful context and turn the text into a non-hateful one. For example, in the first example of Figure \ref{fig:correction_examples}, the VLM can figure out that the use of the word ``black" is directly referring to the man in the visual part, leading to a replacement with "beautiful" instead of "black" and positive promotion of diversity, which is a respectful content for social media. Similar patterns can be observable from other examples about gender and immigration status.

On the other hand, failed examples demonstrate the limitation of VLM in understanding the hidden meaning. In Figure \ref{fig:correction_failed}, the first meme attacks Shia men based on their religious beliefs since it mentions the negative stereotype of Muslims using sensitive terms about sexuality. The VLM attempts to avoid this hatefulness by changing to the new context about enjoying delicious food. However, during the reasoning process, it overlooks the Muslim's food restriction. The second and third examples then reinforce our observation. Our exploration, therefore, calls for additional exploration of whether VLM/LLM could consider causalities to have better reasoning ability.

\section{Conclusion}
While hateful meme detection is well-explored, hateful meme correction is still an open problem. Our work is one of the first to address the hateful memes correction problem. We introduce a simple but efficient approach for detecting and correcting hateful memes by prompting a pretrained VLM. Remarkably, the VLM showcases an impressive ability to extract the content and do the reasoning for requested tasks with only zero-shot prompting. Our quantitative evaluation unfolds promising results achieved without the need for fine-tuning, especially on the hatefulness correction. The use of VLM offers a user-friendly solution through multimodal AI applications. Hence, our exploration calls for more research studies on enhancing the performance of large models on emergent abilities. In the future, we will improve the performance with in-context learning and chain-of-thought techniques.

\bibliographystyle{plainnat}

\appendix
\section{Baseline Methods}
\label{sec:apdix_baselines}
Facebook introduced the Hateful Memes Challenge in late 2020 to not only provide a reference benchmark dataset to tackle the multimodal hate speech detection task but also to provide common visio-linguistic baselines approaches. Indeed, \cite{kiela2020hateful} tested and provided 11 vision and language models through their software MMF \footnote{https://github.com/facebookresearch/mmf}. MMF (MultiModal Framework) is a framework for vision and language multimodal research with state-of-the-art models. In \cite{alayrac2022flamingo, awadalla2023openflamingo}, some preliminary results show the potential of multimodal LLMs in detecting hateful memes with both zero-shot and few-shot learning.

\subsection{Unimodal}
\noindent\textbf{Image-Grid.}
The Image-Grid baseline uses features from ResNet-152's res-5c layer after average pooling (layer of 2048 neurons). ResNet-152 is a Deep Residual Neural Network with 152 layers originally designed for image/scene classification \cite{he2016deep}.

\noindent\textbf{Image-Region.}
The Image-Region baseline uses features from the Faster-RCNN's fc6 layer \cite{ren2015faster} with a ResNeXt-152 \cite{xie2017aggregated} backbone (layer of 4096 neurons). The Fast Region-based Convolutional Neural Network is originally trained on a Visual Genome, and the resulting fc6 features are fine-tuned using weights of the fc7 layer.

\noindent\textbf{Text-BERT.}
The Text BERT baseline is the original BERT model that outputs a vector of dimension 768. BERT is a bidirectional transformer-based model for language representation and understanding that learns embeddings for subwords \cite{devlin2018bert}.

\subsection{Multimodal: unimodal pretraining}
Multimodal models from unimodal pretraining typically combine the output or the features of vision and linguistic models.

\noindent\textbf{Late Fusion.}
In the Late Fusion baseline, the output is the mean of the ResNet-152 and BERT output scores.

\noindent\textbf{Concat BERT.}
The Concat BERT baseline uses the concatenation of the ResNet-152 features with the BERT features, where an MLP is trained on top of it.

\noindent\textbf{MMBT-Grid.}
The MMBT-Grid baseline is the original supervised multimodal bi-transformers (MMBT) \cite{kiela2019supervised} that uses the features of Image-Grid. MMBT is a multimodal bi-transformer model that combines BERT as the textual encoder and ResNet-152 as the image encoder.

\noindent\textbf{MMBT-Region.}
The MMBT-Grid baseline is the original supervised multimodal bi-transformers (MMBT) \cite{kiela2019supervised} that uses the features of Image-Region.

\noindent\textbf{ViLBERT.}
The ViLBERT baseline is the unimodally pretrained version of ViLBERT \cite{lu2019vilbert}.

\noindent\textbf{Visual BERT.}
The Visual BERT baseline is the unimodally pretrained version of Visual BERT \cite{li2019visualbert}.

\subsection{Multimodal: multimodal pretraining}
\noindent\textbf{ViLBERT CC.} The ViLBERT CC baseline is the multimodally pretrained version (also the original version) of ViLBERT that was trained on Conceptual Caption \cite{sharma2018conceptual} for the tasks of sentence-image alignment, masked language modeling, and masked visual-feature classification. ViLBERT is a BERT-based model designed to learn joint contextualized representations of vision and language by using two separate transformers: one for vision and one for language \cite{lu2019vilbert}.

\noindent\textbf{Visual BERT COCO.}
The Visual BERT COCO baseline is the multimodally pretrained version (also the original version) of Visual BERT that was trained on COCO Captions \cite{chen2015microsoft} for the tasks of sentence-image alignment and masked language modeling. [ref to COCO]. Like ViLBERT, it is based on BERT. It was originally designed to learn joint representations of language and visual content from paired data. Unlike ViLBERT, it uses a single cross-modal transformer to align elements of the input text and regions in the input image \cite{li2019visualbert}.

\subsection{Challenge Winner}
After the Hateful Memes Challenge, many state-of-the-art models were proposed to deal with the task. \cite{zhu2020enhance} extracted and then applied demographics while \cite{muennighoff2020vilio} utilized Stochastic Weight Averaging method to stabilize the model. In \cite{velioglu2020detecting}, the authors applied an augmented dataset on original data to train the VisualBERT model. \cite{lippe2020multimodal} used a weighted linear combination of ensemble learners to get the final prediction instead of the majority voting technique. The multimodal deep ensemble technique is leveraged by \cite{sandulescu2020detecting} to boost the performance.

\subsection{Multimodal LLM}
\noindent\textbf{Flamingo (Fl).} This VLM  \cite{alayrac2022flamingo} executes vision-language tasks effectively using pretrained visual encoders and language models. To connect those modules, the Perceiver Resampler is trained to project extensive embedding features from the encoder (NFNet-F6) to visual tokens. Then, both visual and textual tokens are combined and sent to the language model. The method further modifies the pretrain LLM by adding gated cross-attention layers and training those layers from scratch. The cross-attention scheme allows the model to process multiple image-text inputs, opening the ability to perform in-context learning.

\noindent\textbf{OpenFlamingo (OF).} This framework \cite{awadalla2023openflamingo} is an attempt to develop an open-source version of Flamingo and reproduce the results.

\noindent\textbf{LLaVA-Llama-2-13B (LL-2).} Pretrained CLIP and Llama-2 serve as the visual encoder and language model in this VLM \cite{liu2023visual}, respectively. While Flamingo uses the cross-attention scheme, LLaVA proposes a projection layer to map CLIP features to language tokens. Then, both visual and textual tokens are used as inputs to the pretrained language model.

\end{document}